\documentclass[runningheads]{llncs}
\usepackage[T1]{fontenc}
\usepackage{graphicx,verbatim}
\usepackage{xcolor}
\usepackage[dvipsnames]{xcolor}
\definecolor{mygreen}{RGB}{114, 240, 126}
\usepackage{multirow}
\usepackage{amsmath}
\usepackage{amssymb}
\usepackage{booktabs}
\usepackage{pifont}
\usepackage{enumitem}
\usepackage{colortbl}
\usepackage{subcaption}
\usepackage{bbding}
\usepackage{hyperref} 
\usepackage{arydshln}  
\usepackage{tikz}
\usepackage[capitalize]{cleveref}
\crefname{section}{Sec.}{Secs.}
\Crefname{section}{Section}{Sections}
\Crefname{table}{Table}{Tables}
\crefname{table}{Tab.}{Tabs.}

\begin{document}

\title{UltraStar: Semantic-Aware Star Graph Modeling for Echocardiography Navigation}
\titlerunning{Semantic-Aware Star Graph Modeling for Echocardiography Navigation}

\author{
Teng Wang\inst{1}\textsuperscript{*}
\and
Haojun Jiang\inst{1}\textsuperscript{*}\textsuperscript{\ensuremath{\ddag}}
\and
Chenxi Li\inst{1}
\and
Diwen Wang\inst{2}
\and
Yihang Tang\inst{2}
\and
Zhenguo Sun\inst{3}
\and
Yujiao Deng\inst{4}
\and
Shiji Song\inst{1}
\and
Gao Huang\inst{1}\textsuperscript{\ensuremath{\dagger}}
}

\authorrunning{Wang et al.}

\institute{
Department of Automation, BNRist, Tsinghua University, Beijing, China
\and
School of Automation, Beijing Institute of Technology, Beijing, China
\and
Beijing Academy of Artificial Intelligence, Beijing, China
\and
Chinese PLA General Hospital, Beijing, China 
\\
\email{\{t-wang25, jhj20\}@mails.tsinghua.edu.cn, gaohuang@tsinghua.edu.cn}
}

\maketitle               

\begingroup
\renewcommand{\thefootnote}{\fnsymbol{footnote}}
\setcounter{footnote}{0}
\footnotetext[1]{Equal contribution. \quad \(\ddagger\) Guided this work. \quad \(\dagger\) Corresponding author.}
\endgroup

\begin{abstract}

Echocardiography is critical for diagnosing cardiovascular diseases, yet the shortage of skilled sonographers hinders timely patient care, due to high operational difficulties.
Consequently, research on automated probe navigation has significant clinical potential.
To achieve robust navigation, it is essential to leverage historical scanning information, mimicking how experts rely on past feedback to adjust subsequent maneuvers.
Practical scanning is an exploratory trial-and-error process that inherently generates noisy trajectories. However,  
existing methods typically model this history as a sequential chain, forcing models to overfit these noisy paths, leading to performance degradation on long sequences.
In this paper, we propose UltraStar, 
which reformulates probe navigation from path regression to anchor-based global localization.
By establishing a Star Graph, UltraStar treats historical keyframes as spatial anchors connected directly to the current view, explicitly modeling geometric constraints for precise positioning.
We further enhance the Star Graph with a semantic-aware sampling strategy that actively selects the representative landmarks from massive history logs, reducing redundancy for accurate anchoring.
Extensive experiments on a dataset with over 1.31 million samples demonstrate that UltraStar outperforms baselines and scales better with longer input lengths, revealing a more effective topology for history modeling under noisy exploration.
Code is available at https://github.com/LeapLabTHU/UltraStar.

\end{abstract}

\section{Introduction}

\begin{figure}[!t]
\centering
\includegraphics[width=1\columnwidth]{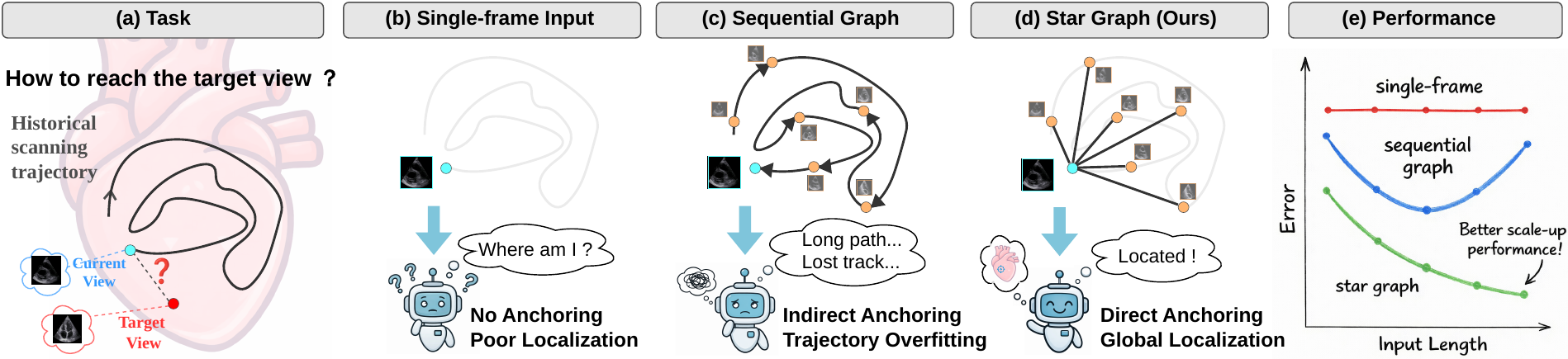}
\caption{
Comparison of modeling paradigms.
(a) The echocardiography navigation task.
(b) Single-frame methods struggle with localization due to limited context.
(c) Sequential Graph methods model history as a chain, forcing the model to overfit noisy exploration trajectories and degrading localization accuracy.
(d) Our Star Graph breaks the chain, treating historical keyframes as global anchors and learning direct geometric constraints for precise localization.
(e) Our method achieves lower error and scales better.
}
\label{fig:motivation}
\end{figure}

Cardiovascular diseases are the leading cause of death~\cite{roth2017global}.
To assess cardiovascular health, echocardiography is an effective method~\cite{mitchell2019guidelines}, being both non-invasive and free from radiation. 
In practice, sonographers need to manually adjust and explore the probe between the ribs, continuously optimizing the angle based on previous exploration trajectory to approach the target view. 
This demanding process requires strong anatomical knowledge and proficient skills, contributing to the scarcity of trained professionals~\cite{won2024sound}.

To address this issue, with the rapid adoption of AI techniques in medical imaging~\cite{christensen2024vision,jiao2024usfm,mh2024lvm,yue2025chexworld,zhang2023biomedclip}, researchers have begun leveraging AI to empower ultrasound probe navigation systems~\cite{bao2024real,chen2025ultradp,droste2020automatic,jiang2025ultrasep,jiang2025towards,narang2021utility,yue2025echoworld}, 
aimed at achieving precise probe navigation to assist sonographers or drive autonomous robotic ultrasound systems.
Several previous studies have demonstrated the potential of AI-driven guidance.
For example, Narang et al.~\cite{narang2021utility} proposed a commercial CNN-based echocardiography navigation system to guide nurses in acquiring standard views, but it is closed-source, limiting broader development. Bao et al.~\cite{bao2024real} proposed a convolutional model for rotational suggestions, but it is limited to A4C and A2C views, reducing its clinical scope. Droste et al.~\cite{droste2020automatic} introduced US-GuideNet, which uses a GRU to model historical image-motion sequences for fetal plane scanning.
Jiang et al.~\cite{jiang2025ultrasep} proposed UltraSeP, which leverages a sequence-aware pre-training paradigm to learn personalized cardiac structures.
Wang et al.~\cite{wang2025ultrahit} developed UltraHiT, employing a hierarchical transformer to model history scanning sequences for autonomous internal carotid artery scanning.
Additionally, Yue et al.~\cite{yue2025echoworld} proposed EchoWorld, a motion-aware world modeling framework that pre-trains a cardiac world model to improve probe guidance.

As mentioned earlier, cardiac ultrasound scanning is typically a process of exploration followed by gradual convergence, where sonographers rely on prior exploration to guide subsequent maneuvers.
Therefore, leveraging historical scanning data is indispensable for robust probe navigation, as single-frame input methods~\cite{jiang2024structure,jiang2024cardiac} lack sufficient context to resolve the ambiguity of the complex cardiac anatomy.
While recent works adopt sequence modeling to capture this context, limitations still remain in the utilization of historical information.
As illustrated in \cref{fig:motivation}, cardiac ultrasound scanning is a complex trial-and-error exploration process.
Consequently, the raw history contains substantial redundant or noisy motions that do not directly aid target acquisition.
However, most existing sequential methods~\cite{droste2020automatic,jiang2025ultrasep,wang2025va,wang2025ultrahit} model the history as a chain (\cref{fig:motivation}(c)). 
This paradigm forces the model to overfit the noisy exploration path, paying excessive attention to step-by-step wandering motions. As a result, the geometric constraints needed for navigation are weakened, and performance often degrades as the sequence length grows.
A key insight of our work is that the main value of history lies not in reconstructing the trajectory, but in global localization—helping the agent understand its current position relative to the anatomy.
To this end, we propose UltraStar, a simple yet effective framework that shifts from chain-based modeling to a Star Graph topology (\cref{fig:motivation}(d)). 
By treating historical keyframes as spatial anchors and directly connecting them to the current view, UltraStar learns the geometric constraints between the current state and historical landmarks, enabling robust localization and superior scalability.
Furthermore, raw scanning history can be excessively long and sparse, making full-trajectory processing computationally infeasible. 
We therefore propose a semantic-aware sampling strategy that selects keyframe landmarks with high semantic divergence, constructing a compact yet informative map for the Star Graph. 
We validate our method on a large-scale dataset of 1.31 million samples, where UltraStar outperforms baselines and demonstrates superior scalability.

\begin{figure}[!t]
\centering
\includegraphics[width=1\columnwidth]{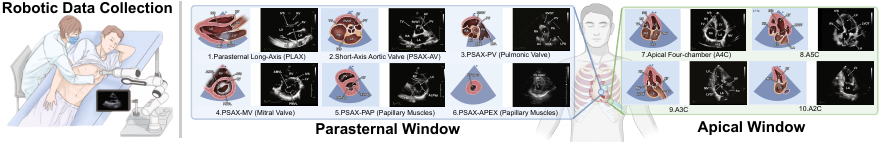}
\caption{
Illustration of the dataset.
The view images are sourced from \cite{mitchell2019guidelines}.
}
\label{fig:dataset}
\end{figure}

\section{Dataset and Method}
In this section, we first describe the echocardiography scanning dataset to provide a background about our task.
Next, we introduce the Star Graph modeling paradigm.
Lastly, we introduce the semantic-aware sampling strategy.

\subsection{Echocardiography Scanning Dataset}
\label{method:dataset}
In this work, we focus on six imaging planes in the parasternal window and four imaging planes in the apical window (\cref{fig:dataset} right).  
To support model training, we built a robotic acquisition system in which a sonographer operates a probe mounted on a robotic arm (\cref{fig:dataset} left), while the probe captures images and the system records its 6-DOF pose.  
The pose is represented by position $(x,y,z)$ and orientation (Euler angles about $x,y,z$).  
Each scan yields a sequence $\{(\mathbf{I}_t, \mathbf{p}_t)\}^{T}_{t=1}$, where $\mathbf{I}_t$ is the image and $\mathbf{p}_t$ is the pose at time $t$, and the relative action $\mathbf{a}_{i \rightarrow j}$ between two poses $\mathbf{p}_{i}$ and $\mathbf{p}_{j}$ can be computed.  
During each scan, sonographers annotate timestamps $\{t_i\}_{i=1}^{10}$ corresponding to the ten standard planes, from which the target poses $\mathbf{p}_{t_i}$ are obtained.  
For any other frame $(\mathbf{I}_{t_j}, \mathbf{p}_{t_j})$ with $j \neq i$, we compute the action $\mathbf{a}_{t_j \rightarrow t_i}$ to the target plane as supervision for training.  
Our dataset was collected by two expert sonographers from 178 adult patients, resulting in 356 scanning trajectories and 1.31M samples in total.

\begin{figure}[!t]
\centering
\includegraphics[width=1\columnwidth]{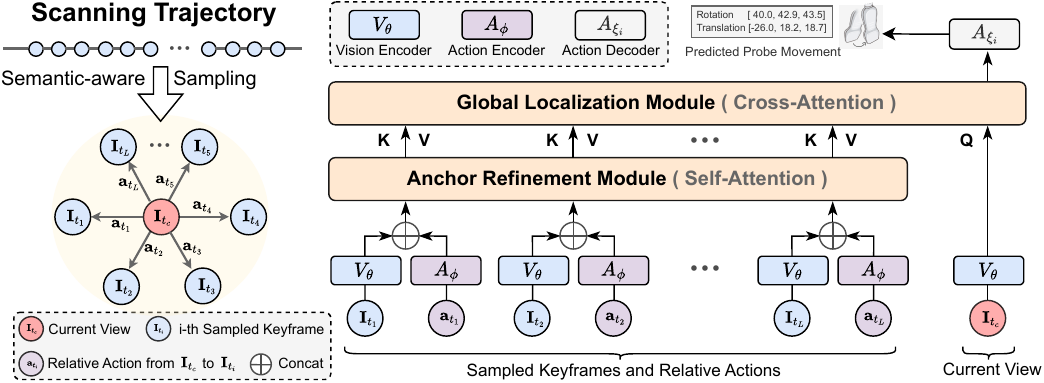}
\caption{
Illustration of the Star Graph modeling paradigm.
}
\label{fig:method}
\end{figure}

\subsection{Star Graph Modeling Paradigm}
\label{method:stargraph}

The core motivation of our method is to utilize historical data for global localization rather than trajectory reconstruction. 
By explicitly modeling the geometric relationship between the current view and historical landmarks, the model can determine its precise location in the cardiac anatomy.
In the following sections, we provide a detailed explanation of our approach, as shown in \cref{fig:method}.

\textbf{Input.}
Given the current image $\mathbf{I}_{t_c}$ from a scan, we first sample $L-1$ historical keyframes $\{\mathbf{I}_{t_1}, \dots, \mathbf{I}_{t_{L-1}}\}$ using the strategy described in \cref{method:sampling}.
Unlike sequential methods that model the path between adjacent historical frames, we construct a Star Graph topology.
We compute the relative actions $\mathbf{a}_{t_c \rightarrow t_i} \in \mathbb{R}^{6}$ from the current view $\mathbf{I}_{t_c}$ to each sampled keyframe $\mathbf{I}_{t_i}$.
This forms a set of spatial anchors, where each anchor consists of a visual appearance and its geometric position relative to the current view:
\begin{align}
    \mathcal{S} = \{ (\mathbf{I}_{t_i}, \mathbf{a}_{t_c \rightarrow t_i}) \mid i = 1, \dots, L-1 \}.
\end{align}

\textbf{Feature Encoding.}
We utilize a shared vision encoder $V_\theta$ (ViT~\cite{dosovitskiy2020image}) to extract visual features.
The current view $\mathbf{I}_{t_c}$ is encoded into a query feature $\mathbf{f}_{c}^{v} \in \mathbb{R}^{C}$.
Similarly, each historical keyframe $\mathbf{I}_{t_i}$ is encoded into a visual feature $\mathbf{f}_{i}^{v} \in \mathbb{R}^{C}$.
Simultaneously, the relative 6-DOF actions are processed by an action encoder $A_\phi$ to map the geometric constraints into the same feature space:
\begin{align}
    \mathbf{f}_{c}^{v} = V_\theta(\mathbf{I}_{t_c}), \ \ \mathbf{f}_{i}^{v} = V_\theta(\mathbf{I}_{t_i}),  \ \
    \mathbf{f}_{i}^{a} = A_\phi(\mathbf{a}_{t_c \rightarrow t_i}), \ \ \mathbf{f} \in \mathbb{R}^{C}.
\end{align}

\textbf{Star Graph Reasoning.}
The reasoning process consists of two stages: Anchor Refinement and Global Localization.
First, to construct robust multimodal anchors, we concatenate the visual and action features for each sampled keyframe.
These raw anchors are then fed into an Anchor Refinement Module (implemented as a two-layer Self-Attention block) to model the correlations among historical landmarks and filter noise:
\begin{align}
    \mathbf{h}_i &= \text{Concat}(\mathbf{f}_{i}^{v}, \mathbf{f}_{i}^{a}) \in \mathbb{R}^{2C}, \\
    [\mathbf{\hat{h}}_1, \dots, \mathbf{\hat{h}}_{L-1}] &= \text{SelfAttn}([\mathbf{h}_1, \dots, \mathbf{h}_{L-1}]).
\end{align}

Next, we employ a Global Localization Module. We project the refined anchors $\hat{\mathbf{h}}$ to dimension $C$ as the Key and Value, and use the current view feature $\mathbf{f}_{c}^{v}$ as the Query for the Cross-Attention mechanism. This operation aggregates geometric evidence from the map to localize the current state:
\begin{align}
    \mathbf{m} &= \text{CrossAttn}(Q=\mathbf{f}_{c}^{v}, K=\mathbf{\hat{h}}, V=\mathbf{\hat{h}}).
\end{align}

\textbf{Prediction and Loss.}
Finally, the localized feature $\mathbf{m}$ is passed to task-specific action decoders $A_{\xi_{k}}$ to predict the relative action $\mathbf{a}_{t_c \rightarrow \text{target}_k}$ required to reach the $k$-th standard plane.
We adopt millimeters for translation and degrees for rotation. 
This unit choice brings both components to the same order of magnitude, allowing us to assign equal weights to translation and rotation errors during optimization.
The model is trained using the Smooth L1 Loss:
\begin{align}
    \mathbf{a}' = A_{\xi_{k}}(\mathbf{m}), \ \
    \mathcal{L} = \mathcal{L}_{\mathrm{SmoothL1}}(\mathbf{a}_{\text{gt}}, \mathbf{a}').
\label{eq:loss}
\end{align}

\subsection{Semantic-aware Sampling Strategy}
\label{method:sampling}

\begin{figure}[!t]
\centering
\includegraphics[width=1\columnwidth]{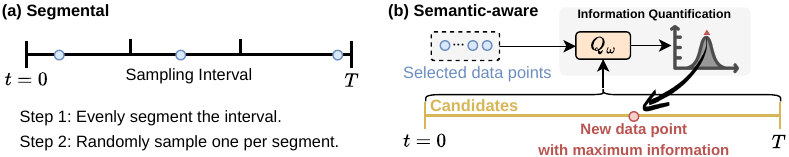}
\caption{
Diagram of segmental sampling and idea of semantic-aware sampling.
}
\label{fig:sampling}
\end{figure}

The efficacy of our UltraStar framework relies on the quality of the spatial anchors used for map construction. 
To achieve precise global localization, the constructed Star Graph must be informationally dense: a set of redundant anchors provides weak geometric constraints, whereas a diverse map facilitates robust triangulation.
Let $L$ denote the number of nodes in the input graph. 
Our goal is to select $L-1$ informative historical anchors from the raw scanning history.

\textbf{Segmental Sampling.}
As a representative baseline, segmental sampling divides the historical trajectory into $L-1$ equal temporal segments and randomly selects one point from each segment (see \cref{fig:sampling}(a)).
This strategy improves temporal coverage compared with unconstrained sampling by encouraging anchors to be distributed across the full exploration process.
However, it remains content-agnostic and does not consider the actual visual information, often resulting in homogeneous anchors that fail to represent the global anatomy effectively.

\textbf{Semantic-aware Strategy.}
To construct an informationally rich map for the Star Graph, we propose a semantic-aware strategy that enhances basic sampling by maximizing the semantic diversity of the anchors (see \cref{fig:sampling}(b)).
Specifically, we train a view classification model $Q_{\omega}$ to quantify the semantic content of each frame. 
For an image $\mathbf{I}$, $Q_{\omega}$ outputs a probability distribution $\mathbf{z} \in\mathbb{R}^{10}$ over 10 standard views.
The sampling process is designed to minimize redundancy.
When selecting a new anchor, we compute its semantic similarity against the current view $\mathbf{I}_t$ and all previously sampled anchors using cosine similarity on their distribution vectors $\mathbf{z}$.
We sum these similarity scores to measure the redundancy of a candidate frame relative to the existing graph nodes.
To ensure diversity while maintaining randomness, we identify the $K$ candidates with the lowest total similarity (i.e., the most distinct frames) and randomly select one as the new anchor.
This iterative process ensures that the resulting Star Graph covers the widest possible semantic range within the cardiac anatomy.

\section{Experiments}

\subsection{Implementation Details}

\textbf{Dataset.}
The dataset was collected using a GE machine (\textit{General Electric}) equipped with a M5S probe.
We use 284 scans for training and 72 scans for validation, with a strict patient-level split.
The validation set comprises individuals not encountered during training.
The vision encoder $V_\theta$ and view classification model $Q_{\omega}$ were trained only on training scans to prevent data leakage.
The data collection was approved by the University Medical Ethics Committee.

\textbf{Model Architecture.}
We use a ViT-Small/16 model as the default vision encoder.
The action encoder is a linear layer that maps the original 6-DOF actions to a 192-dimensional action space to match the vision feature dimension.
The Anchor Refinement Module is implemented as a two-layer Self-Attention block with 4 attention heads.
The Global Localization Module consists of a single Cross-Attention layer with 4 attention heads.
The action decoder is a two-layer MLP with a GELU function in between.
For the view classification model $Q_{\omega}$, we adopt a ResNet-34~\cite{he2016deep} trained on 123K annotated samples.

\textbf{Training Details.}
The model is optimized with AdamW (batch size 128, learning rate $1 \times 10^{-4}$) for 5 epochs using a cosine decay schedule.
The vision encoder $V_\theta$ loads I-JEPA~\cite{assran2023self} parameters pre-trained on our dataset and is frozen during training.
For semantic-aware sampling, the candidate selection hyperparameter $K$ is set to 128. 
All experiments are performed on 8 H20 GPUs.

\textbf{Evaluation Metric.} 
The metric used is the Mean Absolute Error (MAE) between the predicted probe movement action $\mathbf{a}^{'}$ and the ground truth $\mathbf{a_{gt}}$.

\begin{table*}[!t]\scriptsize
\caption{
Comparison with baselines using MAE for \textbf{translation (mm)} and \textbf{rotation ($^\circ$)}.
We report the average MAE over six standard views from Parasternal Window, four standard views from Apical Window (\cref{fig:dataset}), and the overall average across all views.
* indicates $p < 0.0001$.
}
\label{tab1:baselines}
\begin{center}
\resizebox{0.9\columnwidth}{!}{
\begin{tabular}{p{1.6cm} p{2.8cm} *{6}{c}}
\toprule
\multirow{2}{*}{\textbf{\small Type}} & \multirow{2}{*}{\textbf{\small Method}} &
\multicolumn{2}{c}{\textbf{Parasternal}} &
\multicolumn{2}{c}{\textbf{Apical}} &
\multicolumn{2}{c}{\textbf{Average}} \\
\cmidrule(lr){3-4}\cmidrule(lr){5-6}\cmidrule(lr){7-8}
& & Trans. & Rot. & Trans. & Rot. & Trans. & Rot. \\
\midrule

\multirow{7}{*}{\begin{tabular}[c]{@{}l@{}}Single-\\ frame\end{tabular}}

& BioMedCLIP \cite{zhang2023biomedclip}   & 8.24 & 8.55 & 8.75 & 9.76 & \textcolor{red}{8.44} & \textcolor{red}{9.03} \\
& LVM-Med \cite{mh2024lvm}                & 8.35 & 8.46 & 8.86 & 9.56 & \textcolor{red}{8.56} & \textcolor{red}{8.90} \\
& US-MoCo \cite{chen2021empirical}                               & 8.37 & 8.44 & 8.74 & 9.60 & \textcolor{red}{8.51} & \textcolor{red}{8.90} \\
& US-IJEPA \cite{assran2023self}          & 8.13 & 8.20 & 8.67 & 9.38 & \textcolor{red}{8.35} & \textcolor{red}{8.67} \\
& US-MAE \cite{he2022masked}                                  & 8.13 & 8.19 & 8.46 & 9.36 & \textcolor{red}{8.26} & \textcolor{red}{8.66} \\
& USFM \cite{jiao2024usfm}                & 8.12 & 8.19 & 8.36 & 9.27 & \textcolor{red}{8.21} & \textcolor{red}{8.62} \\
& EchoCLIP \cite{christensen2024vision}  & 8.14 & 8.16 & 8.33 & 9.05 & \textcolor{red}{8.21} & \textcolor{red}{8.52} \\

\midrule

& \begin{tabular}[c]{@{}l@{}}GRU \\ (from US-GuideNet~\cite{droste2020automatic})\end{tabular} & 6.97 & 7.20 & 5.09 & 8.26 & \textcolor{red}{6.22} & \textcolor{red}{7.62} \\
\noalign{\vskip 0.8ex}
\arrayrulecolor{gray}\cdashline{2-8}\arrayrulecolor{black}
\noalign{\vskip 0.8ex}

\begin{tabular}[c]{@{}l@{}}Sequential\\ Graph\end{tabular}
& \begin{tabular}[c]{@{}l@{}}Causal self-attn \\ (from Decision-T~\cite{chen2021decision})\end{tabular} & 7.04 & 7.09 & 4.80 & 8.18 & \textcolor{red}{6.14} & \textcolor{red}{7.53} \\
\noalign{\vskip 0.8ex}
\arrayrulecolor{gray}\cdashline{2-8}\arrayrulecolor{black}
\noalign{\vskip 0.8ex}

& \begin{tabular}[c]{@{}l@{}}Non-causal self-attn \\ (from UltraSeP~\cite{jiang2025ultrasep})\end{tabular} & 6.45 & 6.70 & 4.39 & 7.78 & \textcolor{red}{5.63} & \textcolor{red}{7.13} \\

\midrule

\begin{tabular}[c]{@{}l@{}}FC\\ Graph\end{tabular}
& \begin{tabular}[c]{@{}l@{}}Motion-aware attn \\ (from EchoWorld~\cite{yue2025echoworld})\end{tabular} & 5.60 & 6.09 & 4.00 & 7.83 & \textcolor{red}{4.96} & \textcolor{red}{6.79} \\

\midrule

\begin{tabular}[c]{@{}l@{}}\textbf{Star}\\ \textbf{Graph}\end{tabular}
& \begin{tabular}[c]{@{}l@{}}\textbf{UltraStar} \\ \textbf{(Ours)}\end{tabular} & \textbf{5.30} & \textbf{5.67} & \textbf{3.60} & \textbf{7.50} & \begin{tabular}[c]{@{}c@{}}\textcolor{red}{\textbf{4.62}}\\ \textcolor{red}{($\downarrow$7\%)}\end{tabular} & \begin{tabular}[c]{@{}c@{}}\textcolor{red}{\textbf{6.40}}\\ \textcolor{red}{($\downarrow$6\%)}\end{tabular}
\makebox[0pt][l]{
  \hspace{0.15cm} 
  \raisebox{0cm}{
    \begin{tikzpicture}[overlay]
      \draw[black, thin] (0.1, 0.1) -- (0.1, 0.8);
      \draw[black, thin] (0.2, 0.1) -- (0.2, 1.55);
      \draw[black, thin] (0.3, 0.1) -- (0.3, 3.6);
      \node[right, inner sep=1pt, font=\normalsize] at (0.3, 1.8) {*};
    \end{tikzpicture}
  }
}
\\
\bottomrule
\end{tabular}
}
\end{center}
\end{table*}

\subsection{Results and Analysis}
\begin{figure}[t!]
\centering
\includegraphics[width=1\columnwidth]{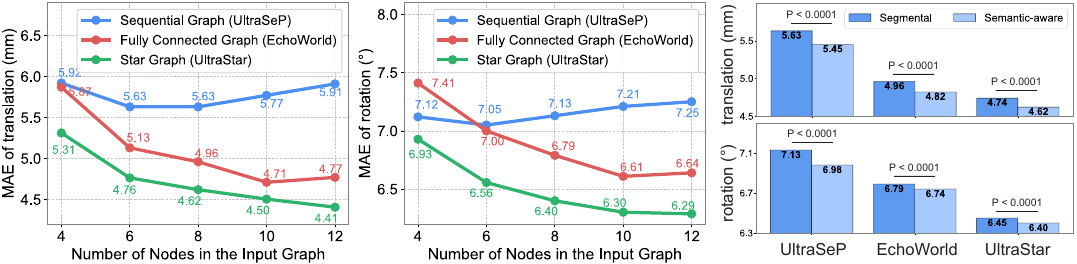}
\caption{
\textbf{Left: Scalability analysis}. As the input graph size increases, our method demonstrates superior scalability with a consistent reduction in translation and rotation errors.
\textbf{Right: Ablation study on sampling strategies}. We compare segmental sampling and semantic-aware sampling across different graph formulations, showing that semantic-aware sampling consistently yields lower translation and rotation errors.
}
\label{fig:input_len}
\end{figure}

\textbf{Comparison with Baseline.}
\Cref{tab1:baselines} presents the quantitative comparison with an input graph size of $L=8$.
To isolate the impact of modeling paradigms, all non-single-frame methods share the same frozen vision encoder.
To ensure a rigorous evaluation, we explicitly exclude frames in the scanning trajectory that are visually similar to the target view from the candidate pool, preventing potential data leakage.
As observed, UltraStar achieves state-of-the-art performance across all metrics.
Sequential Graph methods lag behind because the chain structure encourages overfitting to noisy exploration trajectories.
The Fully Connected Graph also underperforms compared to our method; its dense connectivity introduces excessive noisy edges irrelevant to the current decision, thereby confusing the model.
In contrast, Star Graph establishes direct geometric links between historical anchors and the current view, enabling precise localization.
Furthermore, we analyze the impact of input length in \cref{fig:input_len} (left).
As the input graph size increases, UltraStar demonstrates superior scalability, with a consistent reduction in navigation errors.
This indicates that our method effectively leverages extended historical information to refine localization, whereas baselines saturate or degrade due to the noise introduced by longer trajectories.
We further conduct an ablation study on sampling strategies, as shown in \cref{fig:input_len} (right).
Our semantic-aware sampling strategy consistently yields the lowest errors across all modeling paradigms, confirming its universality and effectiveness in selecting high-quality keyframes.

\begin{figure}[t!]
\centering
\includegraphics[width=1\columnwidth]{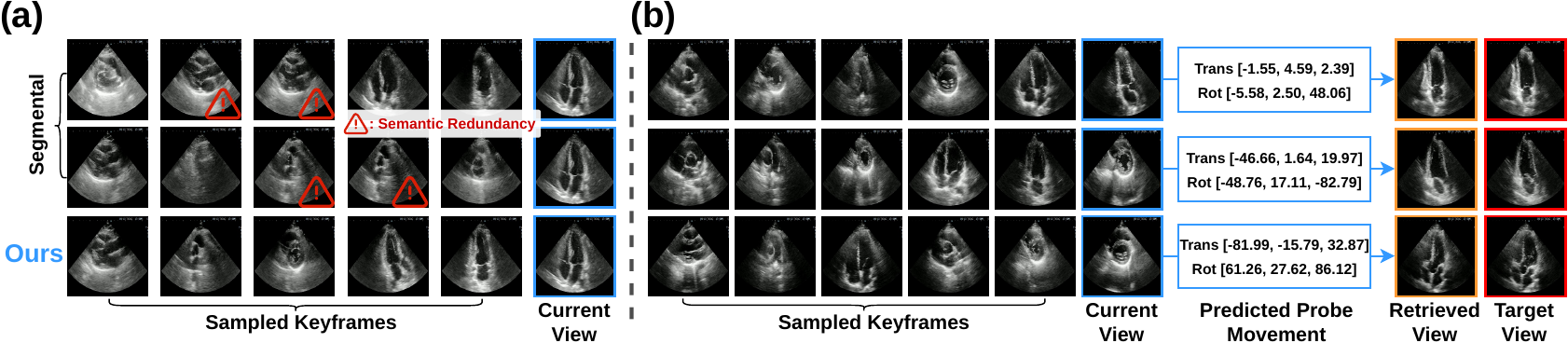}
\caption{
(a) Visualization of the outcome of sampling strategies.
(b) Visualization of the model prediction. Inputs include sampled keyframes, the current view, and the relative probe motions from the current view to each sampled keyframe.
}
\label{fig:vis_sampling}
\end{figure}

\textbf{Visualization.}
First, with the input graph size set to 6, \cref{fig:vis_sampling}(a) illustrates the outcomes of different sampling strategies.
Segmental sampling exhibits distinct semantic redundancy, where multiple sampled keyframes convey repetitive visual information.
In contrast, our semantic-aware strategy selects diverse landmarks, effectively maximizing the information entropy of the Star Graph.
Second, to intuitively demonstrate the model’s output, we calculate the final view pose based on the current view’s pose and predicted probe movement action. 
Using this pose, we perform the nearest-neighbor retrieval from the same scan.
We can observe from \cref{fig:vis_sampling}(b) that our model can output correct probe movement actions, bringing the probe closer to target views.

\section{Conclusion}
In this paper, we presented UltraStar, a framework that reformulates echocardiography probe navigation from path modeling to anchor-based global localization.
By replacing the sequential chain with a Star Graph topology, UltraStar leverages historical frames as spatial anchors for geometric reasoning, establishing direct spatial constraints that ensure precise global localization.
We further introduce a semantic-aware sampling strategy to select representative landmarks, constructing a compact yet informative map from massive scanning logs. Experiments on a large-scale dataset show that UltraStar achieves state-of-the-art performance and scales better with longer history lengths, enabling precise and scalable ultrasound probe navigation.
Beyond echocardiography, this anchor-based paradigm provides a highly effective and novel perspective for general history modeling under unconstrained, noisy exploration. 
To bridge the final gap toward clinical application, our future work will focus on evaluating the framework's performance on real human subjects in real-world settings.

\bibliographystyle{splncs04}
\bibliography{reference}

\end{document}